\documentclass[10pt,twocolumn,letterpaper]{article}

\usepackage{cvpr}              

\usepackage{graphicx}
\usepackage{amsmath}
\usepackage{amssymb}
\usepackage{booktabs}

\usepackage{stackengine,scalerel}
\newcommand\eye{\ensurestackMath{\stackinset{c}{}{c}{-.33pt}%
  {\bullet}{\bigcirc}}}

\usepackage{multirow}
\usepackage[normalem]{ulem}
\useunder{\uline}{\ul}{}
\usepackage[table,xcdraw]{xcolor}

\usepackage[pagebackref,breaklinks,colorlinks]{hyperref}

\usepackage[capitalize]{cleveref}
\crefname{section}{Sec.}{Secs.}
\Crefname{section}{Section}{Sections}
\Crefname{table}{Table}{Tables}
\crefname{table}{Tab.}{Tabs.}

\def\cvprPaperID{35} 
\def\confName{UNCV-W}
\def\confYear{2023}

\DeclareMathOperator{\sigm}{\ensuremath{\text{sigm}}}
\newcommand{\ood}{\ensuremath{\text{OOD}}}
\newcommand{\id}{\ensuremath{\text{ID}}}

\newcommand{\textpresuperscript}[1]{%
\textsuperscript{#1}\nobreak\hspace{0pt}}
  
\begin{document}

\title{On the detection of Out-Of-Distribution samples in Multiple Instance Learning}

\author{Loïc Le Bescond$^{1,2}$
\and
Maria Vakalopoulou$^1$
\and
Stergios Christodoulidis$^1$
\and
Fabrice André$^2$
\and
Hugues Talbot$^1$\\
\textpresuperscript 1CentraleSupélec, \textpresuperscript 2Gustave Roussy \\
{\tt\small name.surname@\{centralesupelec, gustaveroussy\}.fr}
}

\maketitle

\begin{abstract}
The deployment of machine learning solutions in real-world scenarios often involves addressing the challenge of out-of-distribution (OOD) detection. While significant efforts have been devoted to OOD detection in classical supervised settings, the context of weakly supervised learning, particularly the Multiple Instance Learning (MIL) framework, remains under-explored. In this study, we tackle this challenge by adapting post-hoc OOD detection methods to the MIL setting while introducing a novel benchmark specifically designed to assess OOD detection performance in weakly supervised scenarios. Across extensive experiments based on diverse public datasets, KNN emerges as the best-performing method overall. However, it exhibits significant
shortcomings on some datasets, emphasizing the complexity of this under-explored and challenging topic. Our findings shed light on the complex nature of OOD detection under the MIL framework, emphasizing the importance of developing novel, robust, and reliable methods that can generalize effectively in a weakly supervised context. The code for the paper is available here: \href{https://github.com/loic-lb/OOD_MIL}{https://github.com/loic-lb/OOD\_MIL}.

\end{abstract}


\section{Introduction}
\label{sec:intro}

The rapid development of effective machine learning algorithms has facilitated their widespread application across diverse domains, including critical applications such as medical diagnosis~\cite{Sandbank2022, Pantanowitz2020}. Nevertheless, a crucial concern, emphasized by Hendrycks and Gimpel~\cite{Hendrycks2017}, pertains to the challenges faced by machine learning classifiers when deployed in real-world scenarios where the distribution of test and training data differs. Such discrepancies can result in dramatic situations where the model provides inaccurate outputs due to variations in the input arising from different sample collection or preparation protocols. These disparities stem from the assumption made by most machine learning models that all inputs will be drawn from the same distribution used during training process, known as the in-distribution (ID). Consequently, the uncertainty estimation and detection of out-of-distribution (OOD) samples become imperative for the successful application of these algorithms. We distinguish two types of shifts between ID and OOD: the semantic shift, where there is no class overlap between the two distributions, and covariate shift, where the class can overlap, but the style of the input differs. In the context of OOD detection, emphasis is typically placed on the first type~\cite{Yang2021survey}, while the second type is more closely associated with domain generalization~\cite{Koh2021}.

\begin{figure}[t]
  \centering
   \includegraphics[width=0.75\linewidth]{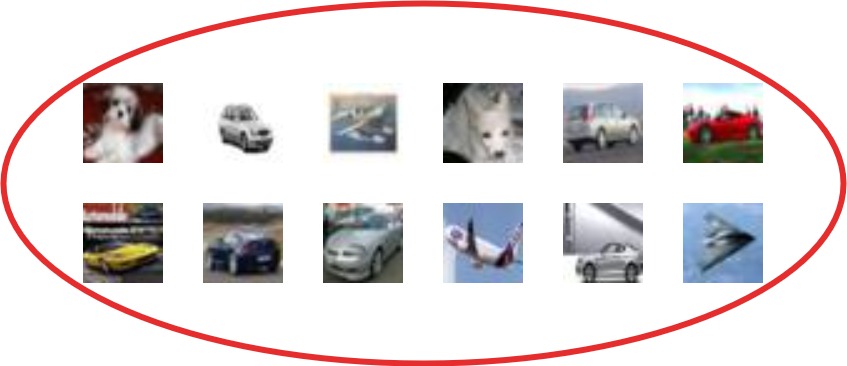}
   \caption{Example of bag sample created from CIFAR10~\cite{Krizhevsky2009}. The positive target class is dogs, and the negative classes are planes and cars. The bag is labeled as positive as it contains two instances of dogs.}
   \label{fig:ex_bag}
\end{figure}

The problem of OOD detection has been extensively addressed in various research works, demonstrating promising performance across different ID datasets and OOD conditions. As described in~\cite{Zhang2023OOD}, these methods can be categorized into three main groups: post-hoc inference methods, which employ pretrained models without further training;  methods that require retraining the model without the use of OOD examples; and methods that necessitate new training with additional OOD data. Among the post-hoc methods, some authors proposed to compute the confidence score directly by considering maximum softmax and logits values derived from the outputs of the penultimate layer of the network~\cite{Hendrycks2017, Hendrycks2022}, or employed energy measures based on the last layer logits~\cite{Liu2020}. Some methods built upon these ideas to improve the confidence score's selectivity~\cite{Liang2017, Sun2022}, while others focused on intermediate features constructing distance-based measures~\cite{Lee2018, Sun2022KNN}. Methods that require retraining often involve modifications to the core architecture, as illustrated by G-ODIN~\cite{Hsu2020}, or the introduction of an additional loss function using OOD data during training~\cite{Hendrycks2019oe}. However, such approaches may hinder the performance of the base classifier and be too specific to the OOD set considered. Therefore, we have opted to focus on post-hoc methods for OOD detection. These methods are particularly appealing with their ease of use, compatibility with various machine learning frameworks, and competitive performance compared to other approaches on multiple benchmarks.

In this study, we propose to adapt post-hoc OOD methods to the Multiple Instance Learning (MIL) framework. Multiple Instance Learning (MIL) has emerged as a dominant approach for weakly supervised image classification. Initially introduced by Dietterich et al.~\cite{Dietterich1997} while exploring drug activity prediction, MIL framework tackles the challenge of classifying a set of input images where individual labels are unknown, with only access to a shared single label for the set. This set of images is commonly referred to as a ``bag,'' and the individual images within it are referred to as ``instances.'' An example of an input bag is illustrated in Fig.~\ref{fig:ex_bag}.  Since its introduction, MIL has found successful applications in diverse medical domains, in particular digital pathology~\cite{Lu2021, Thandiackal2022}. Uncertainty estimation has already been explored to improve Multiple Instance Learning (MIL) approaches. For instance, in~\cite{De2023}, uncertainty estimation is leveraged to identify local artifacts in the input image and enhance MIL generalization. Similarly,~\cite{Schmidt2023} employs Gaussian processes to model the uncertainty of parameters in AttentionMIL~\cite{Ilse2018} to boost the model's performance. Although these approaches have the potential to provide uncertainty estimates for predictions, they do not directly address the problem of OOD detection.

To the best of our knowledge, this is the first attempt to establish a benchmark for OOD detection in the context of MIL. We assert that the assumptions and methodologies underlying traditional post-hoc OOD detection approaches may not be directly applicable to MIL models. The weak supervision context in MIL, where instances are grouped into bags and labeled at the bag level, introduces perturbations into the feature representations and outputs. Consequently, the quality of the embeddings and predictions in the MIL setting may not be as reliable as in the traditional supervised setting. In particular, the main contributions of this work can be summarised as:
\begin{itemize}
    \item We present the first study that evaluates different post-hoc OOD methods in the MIL setting, discussing and comparing their performances.
    \item We use common datasets and organize them in a MIL setting, focusing mainly on semantic shift.
\end{itemize}

All our code and datasets will be made publicly available to help other teams to focus on this challenging and under-explored topic. Our extensive benchmarks highlight the need for more specialized methods for OOD in the MIL settings.

\section{Methods}
\label{sec:method}
\subsection{Multiple Instance Learning}

Consider a classical binary supervised classification problem where the objective is to predict a binary label, denoted as $Y \in \{0,1\}$, based on an input $X$. Within the framework of Multiple Instance Learning (MIL), the variable $X$ represents a collection of instances denoted as $X = \{x_1, ..., x_n\}$, where the individual instance labels $\{y_1, ..., y_n\}$ are unknown. Following the initial formulation by Dietterich et al.~\cite{Dietterich1997}, we assume that $Y$ is positive ($Y=1$) if at least one of the instances $x_i$ has a positive label ($y_i=1$), else $Y$ is considered negative ($Y=0$).

MIL models commonly consist of three main components: an instance embedder $f$, mapping each input $x_i\in\mathbb{R}^{D}$ to a lower-dimensional vector representation $h_i\in\mathbb{R}^M$; a permutation-invariant pooling operator $\theta$, combining all the instance representations extracted from $X$ into a single representation $h\in\mathbb{R}^M$; and a classifier $g$ that generates the final classification score based on the pooled representation. For the image classification task, the instance embedder $f$ typically consists of a CNN architecture, while the classifier $g$ is a simple linear layer. As for the pooling operator $\theta$, we adopt the gated attention mechanism proposed by Ilse et al.~\cite{Ilse2018}. Let $H = \{h_1, ..., h_n\}$ denote  the collection of the representation extracted from $X$ using the embedder $f$. The gated attention pooling mechanism is defined as follows:
\begin{equation}
    h = \sum_{i=1}^{n} a_ih_i
\end{equation}
where, 
\begin{equation}
    a_i = \frac{\exp\{\mathbf{w}^T(\tanh(\mathbf{V}h_i^T)\eye \sigm(\mathbf{U}h_i^T))\} }{\sum\limits_{j=1}^{n}\exp\{\mathbf{w}^T(\tanh(\mathbf{V}h_j^T)\eye \sigm(\mathbf{U}h_j^T))}
\end{equation}
with $\mathbf{w}\in\mathbb{R}^{L\times 1}$, $\mathbf{U}\in\mathbb{R}^{L\times M}$, $\mathbf{V}\in\mathbb{R}^{L\times M}$ trainable parameters, $\sigm$ the non-linear sigmoid activation function and $\eye$ the element-wise (Hadamar) product.

\subsection{Out-of-distribution detection}

As outlined in Zhang et al~\cite{Zhang2023OOD}, the task of out-of-distribution (OOD) detection aims to construct a confidence score that effectively identifies samples $x_{} \sim \mathcal{D}_{\ood}$ while maintaining the model's performance on the in-distribution (ID) dataset $(x_{\id}, y_{\id})\sim\mathcal{D}_{\id}$.

We explored a first set of methods relying solely on the output of the penultimate layer, denoted as $f_c$, namely the maximum softmax probability \textbf{MSP}~\cite{Hendrycks2017}, the maximum logits \textbf{MLS}~\cite{Hendrycks2022} and an energy-based score based on the last layer logits values \textbf{EBO}~\cite{Liu2020}:
\begin{equation}
    C(X^*) = \sigma(f_c(X^*))
\end{equation}
where $X^*$ is a test sample, and $\sigma$ is the composition of the maximum and softmax operators for MSP, the maximum operator for MLS, and the log of summed exponentials for EBO with a temperature scaling parameter $T$. 

In addition to these first methods, we explored further enhancements involving the processing of $f_c$. For instance, we used the recent \textbf{DICE} approach~\cite{Sun2022}, in which certain  weights from $f_c$ are masked based on their relative contribution to enhance the energy score selectivity. Moreover, we investigated the use of input perturbations along temperature scaling with the \textbf{ODIN} method~\cite{Liang2017}, and of distance-based approach with \textbf{KNN}~\cite{Sun2022KNN}. 

In our approach, we propose to associate the penultimate layer $f_c$ to the classifier $g$ of the MIL framework to compute the confidence score directly at the bag level, in contrast to previous works that pool the uncertainty measured over the instances~\cite{Linmans2023}. We also consider the pooled representation $h$ as the feature vector of interest for the KNN method, computing the distance as follows:
\begin{equation}
    C(\bar h^*, k) = ||\bar h^*-\bar h_{(k)}||_2
\end{equation}
where $\bar h^*$ represents the normalized pooled representation of the test sample $X^*$, and $\bar h{(k)}$ is the normalized pooled representation of the $k$-th nearest neighbor.

\subsection{Generation of OOD dataset for MIL}

No standardized benchmark is currently available for addressing the problem of OOD detection under the weakly supervised setting. Taking inspiration from the experiment proposed by Ilse et al.~\cite{Ilse2018}, we designed our own binary weakly supervised task using different common and public databases.

In this setting, an input $X$, referred to as a bag, consists of a variable number of instances randomly sampled from the original database $\mathcal{D}$. Firstly, we define a positive target class based on the original database. If a bag contains at least one instance belonging to the positive target class, the bag is labeled as positive; otherwise, it is labeled as negative. For the negative instances, we established a set of negative classes, encompassing all classes except the positive target class. This allows us to control the difficulty of the classification problem by adjusting the number of negative classes.

\section{Experiments}
\label{sec:exp}
\subsection{Datasets}

We conducted evaluations on several datasets to assess the performance of our proposed approach. Specifically, we employed MNIST as in-distribution (ID) dataset, with Fashion-MNIST~\cite{Xiao2017}, and KMNIST~\cite{Clanuwat2018} as out-of-distribution datasets. Additionally, we explored both CIFAR10~\cite{Krizhevsky2009} and PCAM~\cite{Bejnordi2017,Veeling2018} as alternative in-distribution datasets. PCAM is a representative dataset for a real-world application scenario, which contains patches of lymph node tissues with both healthy and metastatic tissue. For these two datasets, SVHN~\cite{Yuval2011}, Textures \cite{Cimpoi2014} and places365~\cite{Zhou2018} served as out-of-distribution (OOD) datasets.

To generate the training and validation datasets, we created a balanced set of 20,000 bags for training and 4,000 bags for validation for each ID dataset. The bags were of variable length, following a normal distribution $N\sim\mathcal{N}(10, 2)$, and were composed of images uniformly sampled from the corresponding ID dataset's training set. In our experiments, we focused on the digit "5" for the MNIST dataset and dogs for CIFAR10 as the positive target classes. The negative instances included all other digits for the MNIST dataset and images of planes and cars for CIFAR10. For PCAM, we selected the patches containing metastatic tissue as the positive target class and the other patches as negative instances. The number of positive instances in positive bags ranged from 1\% to 40\% of the bag length, sampled uniformly. For the test and OOD datasets, we generated a balanced set of 400 bags under the same conditions. OOD bags contains samples extracted only from the corresponding OOD database.

\subsection{Experimental Setup}

All our models were implemented using PyTorch 2.0~\cite{Paszke2019}. The training was conducted with a learning rate of $5.10^{-5}$ and a weight decay of $10^{-5}$ with a batch size of $1$. For the MNIST-based MIL dataset, we employed the tile embedder proposed by Ilse et al.~\cite{Ilse2018} which consists of 2 convolutional blocks with maxpooling followed by a linear layer. Regarding the CIFAR10-based MIL and PCAM-based MIL datasets, we adopted a similar approach to many previous state-of-the-art MIL methods~\cite{Lu2021, Thandiackal2022},  and replaced the first convolution blocks by a ResNet50 model pre-trained on ImageNet, which remained frozen during the training process. A linear layer was used as the classifier in all experiments, and the gated attention mechanism was identical as well. To enhance the model's generalization capabilities, random augmentations such as rotation, horizontal flips, and vertical flips were applied to the instances composing the training bags.

\begin{table*}[t]
\centering
\begin{tabular}{|c|c|c|cccccc|}
\hline

\multirow{2}{*}{\textit{ID dataset}} & \multirow{2}{*}{OOD dataset} & \multirow{2}{*}{Metric} & \multicolumn{6}{c|}{Method} \\ \cline{4-9}
& &  & MSP~\cite{Hendrycks2017}                          & MLS~\cite{Hendrycks2022}                                    & EBO~\cite{Liu2020}                                    & ODIN~\cite{Liang2017}                                   & DICE~\cite{Sun2022}                             & KNN~\cite{Sun2022KNN}                          \\ \hline

                                                             &                                 & \cellcolor[HTML]{EFEFEF}AUC$\uparrow$ & \cellcolor[HTML]{EFEFEF}92.33 & \cellcolor[HTML]{EFEFEF}91.83        & \cellcolor[HTML]{EFEFEF}91.77         & \cellcolor[HTML]{EFEFEF}{\ul 94.13} & \cellcolor[HTML]{EFEFEF}\textbf{99.05}   & \cellcolor[HTML]{EFEFEF}62.33          \\
                                                             & \multirow{-2}{*}{\begin{tabular}[c]{@{}c@{}}Fashion-\\ MNIST\end{tabular}} & FPR@95$\downarrow$                    & 65.75                         & 67.75                                 & 69.50                                  & {\ul 46.50}                          & \textbf{02.25}                            & 90.00                                  \\ \cline{2-9} 
                                                             &                                 & \cellcolor[HTML]{EFEFEF}AUC$\uparrow$ & \cellcolor[HTML]{EFEFEF}{\ul 84.74} & \cellcolor[HTML]{EFEFEF}84.54          & \cellcolor[HTML]{EFEFEF}84.49    & \cellcolor[HTML]{EFEFEF}83.38 & \cellcolor[HTML]{EFEFEF}64.42          & \cellcolor[HTML]{EFEFEF}\textbf{86.54}         \\
\multirow{-4}{*}{\textit{MNIST}}                         & \multirow{-2}{*}{KMNIST}        & FPR@95$\downarrow$                    &{\ul 64.25}                        &64.75                            &65.00                                  &69.25                        & 90.50                                  & \textbf{50.00}                                  \\ \hline
\multicolumn{1}{|r|}{}                                       &                                 & \cellcolor[HTML]{EFEFEF}AUC$\uparrow$ & \cellcolor[HTML]{EFEFEF}70.13 & \cellcolor[HTML]{EFEFEF}{\ul 70.98} & \cellcolor[HTML]{EFEFEF}\textbf{71.01} & \cellcolor[HTML]{EFEFEF}54.75          & \cellcolor[HTML]{EFEFEF}48.37          & \cellcolor[HTML]{EFEFEF}64.67         \\
\multicolumn{1}{|r|}{}                                       & \multirow{-2}{*}{places365}     & FPR@95$\downarrow$                    & {\ul 78.25}                   & {\ul 78.25}                            & \textbf{75.50}                         & 94.50                                  & 99.50                                  & 95.00                                  \\ \cline{2-9} 
\multicolumn{1}{|r|}{}                                       &                                 & \cellcolor[HTML]{EFEFEF}AUC$\uparrow$ & \cellcolor[HTML]{EFEFEF}45.65 & \cellcolor[HTML]{EFEFEF}48.46          & \cellcolor[HTML]{EFEFEF}{\ul 48.52}    & \cellcolor[HTML]{EFEFEF}40.37          & \cellcolor[HTML]{EFEFEF}41.81          & \cellcolor[HTML]{EFEFEF}\textbf{94.63} \\
\multicolumn{1}{|r|}{}                                       & \multirow{-2}{*}{SVHN}          & FPR@95$\downarrow$                    & 97.50                         & 96.75                                  & {\ul 96.00}                            & 99.50                                  & 99.75                                  & \textbf{37.25}                         \\ \cline{2-9} 
\multicolumn{1}{|r|}{}                                       &                                 & \cellcolor[HTML]{EFEFEF}AUC$\uparrow$ & \cellcolor[HTML]{EFEFEF}49.51 & \cellcolor[HTML]{EFEFEF}51.51          & \cellcolor[HTML]{EFEFEF}{\ul 51.59}    & \cellcolor[HTML]{EFEFEF}41.33          & \cellcolor[HTML]{EFEFEF}36.65          & \cellcolor[HTML]{EFEFEF}\textbf{91.51} \\
\multicolumn{1}{|r|}{\multirow{-6}{*}{\textit{CIFAR10}}} & \multirow{-2}{*}{Textures}      & FPR@95$\downarrow$                    & 91.00                         & 89.75                                  & {\ul 87.50}                            & 97.00                                  & 99.25                                  & \textbf{45.50}                         \\ \hline
                                                             &                                 & \cellcolor[HTML]{EFEFEF}AUC$\uparrow$ & \cellcolor[HTML]{EFEFEF}30.27 & \cellcolor[HTML]{EFEFEF}34.37          & \cellcolor[HTML]{EFEFEF}35.38          & \cellcolor[HTML]{EFEFEF}51.08    & \cellcolor[HTML]{EFEFEF}{\ul 78.23} & \cellcolor[HTML]{EFEFEF}\textbf{92.62}          \\
                                                             & \multirow{-2}{*}{places365}     & FPR@95$\downarrow$                    & 100.00                         & 100.00                                 & 100.00                                 & 100.00                                 & {\ul 71.25}                         & \textbf{37.00}                            \\ \cline{2-9} 
                                                             &                                 & \cellcolor[HTML]{EFEFEF}AUC$\uparrow$ & \cellcolor[HTML]{EFEFEF}16.48 & \cellcolor[HTML]{EFEFEF}  18.51        & \cellcolor[HTML]{EFEFEF}18.86          & \cellcolor[HTML]{EFEFEF}50.55 & \cellcolor[HTML]{EFEFEF}{\ul 68.54}    & \cellcolor[HTML]{EFEFEF}\textbf{99.04}          \\
                                                             & \multirow{-2}{*}{SVHN}          & FPR@95$\downarrow$                    & 100.00                        & 100.00                                 & 100.00                                 & 100.00                        & {\ul 89.50}                                  & \textbf{01.75}                            \\ \cline{2-9} 
                                                             &                                 & \cellcolor[HTML]{EFEFEF}AUC$\uparrow$ & \cellcolor[HTML]{EFEFEF}37.09 & \cellcolor[HTML]{EFEFEF}42.43          & \cellcolor[HTML]{EFEFEF}44.43          & \cellcolor[HTML]{EFEFEF}{\ul 57.21}    & \cellcolor[HTML]{EFEFEF}56.02 & \cellcolor[HTML]{EFEFEF}\textbf{98.23}          \\
\multirow{-6}{*}{\textit{PCAM}}                          & \multirow{-2}{*}{Textures}      & FPR@95$\downarrow$                    & 97.50                         & 95.75                                  & 94.50                                 &  98.00                           & {\ul 93.00}                         &\textbf{06.25}                                  \\ \hline
\end{tabular}
\caption{Results of the different OOD detection methods for a MIL model trained on a bag version of the MNIST, CIFAR10~\cite{Krizhevsky2009} and PCAM~\cite{Bejnordi2017,Veeling2018} datasets.  The accuracy of the model for the in-distribution (ID) task is 94.75, 90.50 and 78.25 for MNIST, CIFAR10~\cite{Krizhevsky2009} and PCAM~\cite{Bejnordi2017,Veeling2018}, respectively. \textbf{Bold} denotes best performance, and \underline{underline} denotes the second best for each metric.}
\label{tab:results}
\end{table*}

During inference on the OOD datasets, the instances were resized to the same dimension as the samples in the training datasets. For DICE and ODIN, we considered the embeddings of non-augmented instances as we found superior performance compared to utilizing the exact augmented instances employed during training. We set the hyperparameters for the OOD methods to the same values as reported in the experiments of the corresponding papers. Performance evaluation was conducted using the common OOD detection metrics, including AUCROC (Area Under the Receiver Operating Characteristic curve) and FPR@95\% TPR (False Positive Rate at 95\% True Positive Rate). These metrics were used to assess the model's ability to detect instances from OOD datasets effectively.

\subsection{Results}

Table.~\ref{tab:results} presents the OOD detection performances for the different methods for a model trained in the context of Multiple Instance Learning. In the MNIST ID experiments, DICE and KNN display the best performance, closely followed by ODIN in the former case and MSP in the latter, for Fashion-MNIST and KMNIST OOD datasets, respectively. However, the performance of DICE and KNN diminishes considerably when evaluated on the other OOD dataset. In contrast, the other methods demonstrate more consistent results. This would indicate that methods relying on the training data may necessitate specific tuning and exhibit limitations in terms of generalization within the context of Multiple Instance Learning. Regarding the FPR@95 results, they are prohibitively high in most cases, with the exception of DICE for Fashion-MNIST OOD.

In the case of CIFAR10 ID and PCAM ID datasets,  KNN consistently demonstrates superiority over the other methods. KNN outperforms the benchmark methods across all OOD datasets, except for places365 OOD in the case of CIFAR10 ID, where EBO performs the best.  As the second-best performing method, DICE performs well with PCAM ID, but its performance lags behind on CIFAR10 ID, where EBO and MSP show better results. In all experiments, except for places365 OOD with CIFAR10 ID, methods relying on classifier outputs and their enhancements consistently exhibit lower performance. FPR@95 remains high in all experiments except for KNN when evaluated on PCAM ID.

These results suggest that KNN appears to be the most reliable method, demonstrating strong performance across most experiments. However, its performance on the places365 OOD dataset with CIFAR10 ID and Fashion-MNIST OOD is notably low, and the improvement is small for KMNIST OOD compared to the other methods, making it challenging to confirm this advantage.  FPR@95 also remains prohibitively high in the case of MNIST and CIFAR10 ID. 

As a general trend, it is worth noting that methods relying on the pooled representation rather than classifier activations appear to perform better when methods based on the final activation outputs perform poorly, and vice versa. Additionally, in the CIFAR10/PCAM experiments, the instance embedder $f$ was fixed, whereas, in the case of MNIST, it was trained alongside the rest of the network. This observation suggests that the approach used to create embeddings for each instance, and consequently the pooled representation, should guide the decision on whether to rely more on intermediate features or classifier outputs for OOD detection.

\section{Conclusion}
\label{sec:concl}

In this study, we present the first benchmark for out-of-distribution (OOD) detection in the context of Multiple Instance Learning (MIL). Through extensive experimentations on various datasets, we have found that methods based on intermediate features, such as KNN, demonstrate strong performance in the context of Multiple Instance Learning. However, the performance is not consistent in every scenario, depending on the specific dataset characteristics and the configurations of MIL models. This lack of robustness of current OOD detection methods points out the need for innovative techniques that can take into consideration characteristics of MIL models, such as the pooling operator, which represents an interesting avenue for future research.

\section*{Acknowledgments}

This work was partially supported by the ANR project Hagnodice ANR-21-CE45-0007 and the PRISM project funded by France 2030 and grant number ANR-18-IBHU-0002.
{\small
\bibliographystyle{ieee_fullname}
\bibliography{egbib}
}

\end{document}